\documentclass[runningheads]{llncs}
\usepackage[english]{babel}
\usepackage[utf8]{inputenc}
\usepackage{graphicx, amsmath, amssymb}
\usepackage{algorithm}
\usepackage[noend]{algpseudocode}
\usepackage{booktabs, tabularx, subfig}
\usepackage{todonotes}
\begin{document}
\title{Explicit topological priors for deep-learning based image segmentation using persistent homology}
\titlerunning{Topological priors for image segmentation}
\author{James R. Clough%\orcidID{0000-1111-2222-3333} 
\and
Ilkay Oksuz%\orcidID{0000-0001-6478-0534} 
\and
Nicholas Byrne%\orcidID{1111-2222-3333-4444}  
\and \\
Julia A. Schnabel%\orcidID{2222--3333-4444-5555} 
\and 
Andrew P. King%\orcidID{1111-2222-3333-4444} 
\thanks{This work was supported by an EPSRC programme Grant (EP/P001009/1) and the Wellcome EPSRC Centre for Medical Engineering at School of Biomedical Engineering and Imaging Sciences, King’s College London (WT 203148/Z/16/Z).
This research has been conducted using the UK Biobank Resource under Application Number 40119.
We would like to thank Nvidia for kindly donating the Quadro P6000 GPU used in this research.
}}
\authorrunning{J. R. Clough et al.}
\institute{School of Biomedical Engineering \& Imaging Sciences, King\rq{}s College London, UK
\email{james.clough@kcl.ac.uk}\\
%\url{http://www.springer.com/gp/computer-science/lncs}
}
\maketitle              % typeset the header of the contribution
\begin{abstract}
We present a novel method to explicitly incorporate topological prior knowledge into deep learning based segmentation, which is, to our knowledge, the first work to do so.
Our method uses the concept of persistent homology, a tool from topological data analysis, to capture high-level topological characteristics of segmentation results in a way which is differentiable with respect to the pixelwise probability of being assigned to a given class.
The topological prior knowledge consists of the sequence of desired Betti numbers of the segmentation.
As a proof-of-concept we demonstrate our approach by applying it to the problem of left-ventricle segmentation of cardiac MR images of 500 subjects from the UK Biobank dataset, where we show that it improves segmentation performance in terms of topological correctness without sacrificing pixelwise accuracy. 
\end{abstract}
\section{Introduction}
\label{sec:intro}
Image segmentation, the task of assigning a class label to each pixel in an image, is a key problem in computer vision and medical image analysis.
%In many cases local pixel intensity information is sufficient for accurate classification.
%Often though, structures in an image are hard to identify without wider context and so global information is required.
The most successful segmentation algorithms now use deep convolutional neural networks (CNN), with recent progress made in combining fine-grained local features with coarse-grained global features, such as in the popular U-net architecture~\cite{Ronneberger2015}. 
Such methods allow information from a large spatial neighbourhood to be used in classifying each pixel.
However, the loss function is usually one which considers each pixel individually rather than considering higher-level structures collectively. 
%
\iffalse
\begin{figure}[ht]
    \centering
    \includegraphics[width=\textwidth]{topo_grad_lv_seg_example_labelled.png}
    \caption{In the task of left ventricle segmentation we have prior knowledge that the myocardium forms a topological circle.
    By incorporating this information into the segmentation task our method produces topologically correct segmentations.}
    % Dice - 74 -> 77 
    % Paths - 71 -> 85
    \label{fig:topograd_example}
    \vspace{-4mm}
\end{figure}\fi
%
\begin{figure}[ht]%
\noindent\makebox[\textwidth]{
    \centering
    \subfloat[]{{\includegraphics[width=0.17\textwidth]{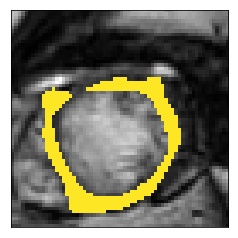}}}%
    \subfloat[]{{\includegraphics[width=0.17\textwidth]{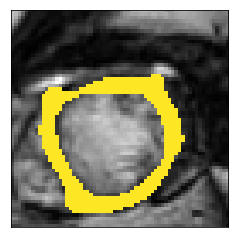}}}%
    \subfloat[]{{\includegraphics[width=0.17\textwidth]{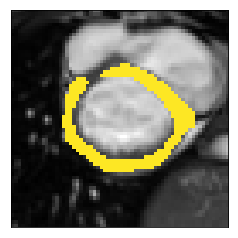}}}%
    \subfloat[]{{\includegraphics[width=0.17\textwidth]{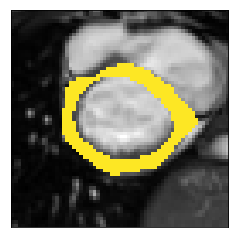}}}%
    \subfloat[]{{\includegraphics[width=0.17\textwidth]{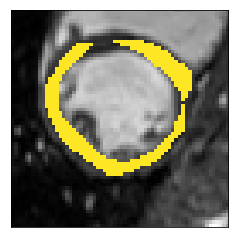}}}%
    \subfloat[]{{\includegraphics[width=0.17\textwidth]{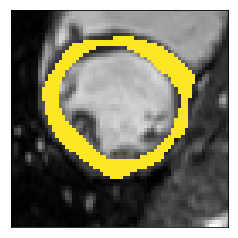}}}}%
    \caption{Three typical clinically acquired MRI images of the short-axis view of the heart.
    The estimated segmentations produced by a U-net model trained with $N_s=100$ supervised cases (a, c, e) show topological errors.
    The segmentations for the same model trained with our topological prior (b, d, f) have the correct topology.}%
    \label{fig:clinical_exemplar}%
\vspace{-3mm}
\end{figure}
%
%
\iffalse
\begin{figure}[ht]%
\noindent\makebox[\textwidth]{
    \centering
    \subfloat[]{{\includegraphics[width=0.19\textwidth]{clinical_example_2/a.png}}}%
    \subfloat[]{{\includegraphics[width=0.19\textwidth]{clinical_example_2/c.png}}}%
    \subfloat[]{{\includegraphics[width=0.19\textwidth]{clinical_example_2/e.png}}}%
    \subfloat[]{{\includegraphics[width=0.19\textwidth]{clinical_example_3/a.png}}}
    \subfloat[]{{\includegraphics[width=0.19\textwidth]{clinical_example_3/c.png}}}%
    \subfloat[]{{\includegraphics[width=0.19\textwidth]{clinical_example_3/e.png}}}}%
    \caption{Two typical clinically acquired MRI images of the short-axis view of the heart (a,d).
    The estimated segmentations produced by a model trained with $N_s=100$ supervised cases (b, e) show topological errors.
    The segmentations for the same model trained with our topological prior (c, f) have the correct topology.}%
    \label{fig:clinical_exemplar}%
\end{figure}
\fi

In many applications it is important to correctly capture the topological characteristics of the anatomy in a segmentation result. 
For example, detecting and counting distinct cells in electron microscopy images requires that neighbouring cells are correctly distinguished.
Even very small pixelwise errors, such as incorrectly labelling one pixel in a thin boundary between cells, can cause two distinct cells to appear to merge.
In this way significant topological errors can be caused by small pixelwise errors that have little effect on the loss function during training but may have large effects on downstream tasks.
Another example is the modelling of blood flow in vessels, which requires accurate determination of vessel connectivity. 
In this case, small pixelwise errors can have a significant impact on the subsequent modelling task.
Finally, when imaging subjects who may have congenital heart defects, the presence or absence of small holes in the walls between two chambers is diagnostically important and can be identified from images, but using current techniques it is difficult to incorporate this relevant information into a segmentation algorithm.
For downstream tasks it is important that these holes are correctly segmented but they are frequently missed by current segmentation algorithms as they are insufficiently penalised during training.
See Figure \ref{fig:clinical_exemplar} for examples of topologically correct and incorrect segmentations of cardiac magnetic resonance images (MRI).

There has been some recent interest in introducing topological features into the training of CNNs, and this literature is reviewed in section \ref{sec:related} below.
However, such approaches have generally involved detecting the presence or absence of topological features \emph{implicitly} in order to quantify them in a differentiable way that can be incorporated into the training of the segmentation network.
The weakness of this approach is that it is hard to know exactly which topological features are being learned.
Instead, it would be desirable to explicitly specify the presence or absence of certain topological features directly in a loss function.
This would enable us to designate, for example, that the segmentation result should have one connected component which has one hole in it. 
This is challenging due to the inherently discrete nature of topological features, making it hard to create a differentiable loss function which accounts for them.

In this paper we demonstrate that persistent homology (PH), a tool from the field of topological data analysis, can be used to address this problem by quantifying the persistence, or stability, of all topological features present in an image.
Our method uses these high-level structural features to provide a pixelwise gradient that increases or decreases the persistence of desired or undesired topological features in a segmentation.
These gradients can then be back-propagated through the weights of any segmentation network and combined with any other pixelwise loss function.
In this way, the desired topological features of a segmentation can be used to help train a network even in the absence of a ground truth, and without the need for those features to be implicitly learned from a large amount of training data, which is not always available.
This topologically driven gradient can be incorporated into supervised learning or used in a semi-supervised learning scenario, which is our focus here.

Our main contribution is the presentation of, to the best of our knowledge, the first method to explicitly incorporate topological prior information into deep-learning based segmentation.
The explicit topological prior is the sequence of desired Betti numbers of the segmentations and our method provides a gradient calculated such that the network learns to produce segmentations with the correct topology.
We begin by reviewing literature related to introducing topology into deep learning in section \ref{sec:related}.
In section \ref{sec:theory} we cover the theory of PH and introduce the relevant notation.
In section \ref{sec:method} we then describe in detail our approach for integrating PH and deep learning for image segmentation, and then demonstrate the method in a case study on cardiac MRI in section \ref{sec:application}.
\section{Related Work}
\label{sec:related}
% Sections to include (and differentiate ourselves from) in this section
% 1- Topological (or allegedly topological) features in image segmentation
%\subsection{Topologically aware segmentation}
The need for topologically aware methods for image processing is becoming increasingly recognised in the literature.
In \cite{Mosinska2017} the task of detecting curvilinear structures was addressed by supplementing a conventional U-net \cite{Ronneberger2015} architecture with a secondary loss function designed to capture topological features.
This was calculated by passing both the ground truth and the predicted segmentations through a pre-trained VGG network \cite{Simonyan2014} and comparing the feature maps at intermediate layers.
These feature maps appeared to capture some topological features such as the presence of small connected components and including this loss function improved performance in the task at hand.
However, it is unclear exactly which topological features were relevant, and how the presence or absence of particular structures were weighted, since they were only captured by the distributed and hard-to-interpret representation of activations in the hidden layers.
Depending on the dataset on which the VGG network was trained, there may be important topological features which were ignored entirely and non-topological features may also contribute to this loss function.
This loss function also still requires ground truth masks, and so cannot be used in a semi-supervised context.
The work of \cite{Oktay2018} used a similar approach in that the output of a second network, in this case an autoencoder, was used to define a loss function which identified global structural features, in order to enforce anatomical constraints, but again, this can only implicitly match the desired topological features and still requires a ground truth segmentation for comparison.

Other approaches have involved encouraging the correct adjacencies of various object classes, whether they were learned from the data as in \cite{Funke2015} or provided as a prior as in \cite{Ganaye2018}.
Such methods allow the introduction of this simple topological feature into a loss function when performing image segmentation but cannot be easily generalised to any other kinds of higher-order feature such as the presence of holes, handles or voids.

% 2- Toporeg paper (topological features of a classifier boundary)
%\subsection{Topological regularisers}
The recent work of \cite{Chen2018b} introduced a topological regulariser for classification problems by considering the stability of connected components of the classification boundary and can be extended to higher-order topological features.
It also provided a differentiable loss function which can be incorporated in the training of a neural network.
This approach differs from ours in that firstly, it imposes topological constraints on the shape of the classification boundary in the feature space of inputs to the network, rather than topological constraints in the space of the pixels in the image, and secondly it aims only to reduce overall topological complexity.
%, encouraging the network to remove complex topological features.
Our approach aims to fit the desired absence or presence of certain features and so complex features can be penalised or rewarded, as is appropriate for the task at hand.

% 3- Other applications of PH to image segmentation, but on the image itself rather than the masks
%    important to explain precisely how we're different here - 
%\subsection{Persistent homology in segmentation problems}
Persistent homology has previously been applied to the problem of semantic segmentation, such as in \cite{Gao2013,Qaiser2016,Assaf2016}.
The important distinction between our method and these previous works is that they apply PH to the input image to extract features, which are then used as inputs to some other algorithm for training.
Such approaches can capture complex features of the input images but require those topological features to be directly extractable from the raw image data.
Our approach instead processes the image with a CNN and it is the output of the CNN, representing the pixelwise likelihood of the structure we want to segment, which has PH applied to it.
%This allows the loss determining the success or failure of the detection of topological features to be back-propagated through the network, encouraging the network to successfully identify them in future.

\section{Theory}
\label{sec:theory}
% Put in the basic theory of persistent homology here
% Perhaps have an example PH for a point cloud if there is space
PH is an algebraic tool developed as part of the growing mathematical field of topological data analysis, which involves computing topological features of shapes and data.
We give a brief overview of PH here, but direct the reader to \cite{Edelsbrunner2002,Edelsbrunner2008,Otter2017} for more thorough reviews, discussions and historical background of the subject.
Although PH most commonly considers \emph{simplicial complexes}%
\footnote{A set of simplices, i.e. points, lines, triangles, tetrahedra, and their higher-dimensional equivalents.
} 
due to their generality, for the analysis of images and volumes consisting of pixels and voxels, \emph{cubical complexes} are considerably more convenient and so we introduce them, and their theory of PH here.
% JC: I think *their theory of PH* rather then *the theory of PH* here, in the sense that different complexes have different ways of computing the PH

\subsection{Cubical Complexes}
% Depending on space we can put some definitions on their own lines, or inline
% Here we introduce the theory of cubical complexes and their notation.
A cubical complex is a set consisting of points, unit line segments, and unit squares, cubes, hypercubes, and so on.
Following the notation of \cite{Kaczynski2006} % don't worry, it's not that T. Kaczynski
its fundamental building blocks are \emph{elementary intervals} which are each a closed subset of the real line of the form 
$I = [z, z+1]$
for 
$z \in \mathbb{Z}$.
These represent unit line segments.
Points are represented by \emph{degenerate} intervals 
$I = [z,z]$.
From these, we can define \emph{elementary cubes}, $Q$ as the %
%finite 
product of elementary intervals,
%i.e. 
\begin{equation}
    Q = I_1 \times I_2 \times ... \times I_d, \quad Q \subset \mathbb{R}^d.
\end{equation}
The set of all elementary cubes in 
$\mathbb{R}^d$ is 
$\mathcal{K}^d$, and 
$\mathcal{K}=\bigcup_d \mathcal{K}^d$.
The dimension of an elementary cube, 
$\mathrm{dim} \, Q$,
is the number of non-degenerate components in the product defining $Q$, and we will denote the set of all d-dimensional elementary cubes as 
$\mathcal{K}_d = \lbrace Q \in \mathcal{K} \, | \, \mathrm{dim} \, Q = d \rbrace $.
By setting up the theory in this way, we are restricting the class of objects we can talk about to unit cubes, which will represent pixels in the images we will consider.
For simplicity, from hereon we will describe the two-dimensional case, but our approach is generalisable.% to higher dimensions.

Consider a 2D array, $S$ of $N \times N$ pixels, where the pixel in row $i$ and column $j$ has a value
$S_{[i,j]} \in [0,1]$.
In terms of the cubical complex each pixel covers a unit square described by the elementary cube
$Q_{i,j}=[i,i+1] \times [j,j+1]$ where $Q_{i,j} \in \mathcal{K}_2$.
We then consider filtrations of this cubical complex.
For each value of a threshold $p \in [0,1]$, we can find the cubical complex $B(p)$ given by
\begin{equation}
    B(p) = \bigcup_{i,j=0}^{N-1} Q_{i,j} %\in \mathcal{K}_2 \, 
    %\mathrm{\ such \ that\ }
    %\mathrm{\ where\ }
    :
    \, S_{[i,j]} \geq (1-p).
\end{equation}
In the context of an image, $B(p)$ represents a binarised image made by setting pixels with a value above $1-p$ to $1$, and below $1-p$ to $0$.
By considering these sets for an increasing sequence of filtration values we obtain a sequence
\begin{equation}
    B(0) \subseteq B(p_1) \subseteq B(p_2) \subseteq ... \subseteq B(1).
\end{equation}
This filtered space is the key object of PH.

\subsection{Persistent Homology}
PH measures the lifetimes of topological features within a filtration such as the sequence above.
The premise is that those features with long lifetimes, in terms of the filtration value $p$, are significant features of the data.
Those with short lifetimes are usually considered to be noise% 
%(although this is not to say that they are always irrelevant)
.
For each complex $B(p) \subset \mathbb{R}^d$, we can consider its topology by finding%
\footnote{We avoid the details of how the homology groups and Betti numbers are computed here.
In our experiments, we used the implementation from the Python library Gudhi, available at \cite{gudhi}.
In our implementation diagonally adjacent pixels are considered as neighbouring, but this does not generally need to be the case.}
the homology group $\mathcal{H}_n$, the rank of which is the n$^{th}$ Betti number, $\beta_n$.
These numbers are topological invariants which, informally speaking, count the number of d-dimensional holes in an object.
$\beta_0$ counts the number of connected components,
$\beta_1$ counts the number of loops, and, although not relevant to the 2D case we consider here, $\beta_2$ counts the number of hollow cavities, and so on.
As the filtration value $p$ increases, more pixels join the cubical complex and topological features in the binarised image are created and destroyed.
\begin{figure}[t!]
\noindent\makebox[\textwidth]{%
\includegraphics[width=1.1\textwidth]{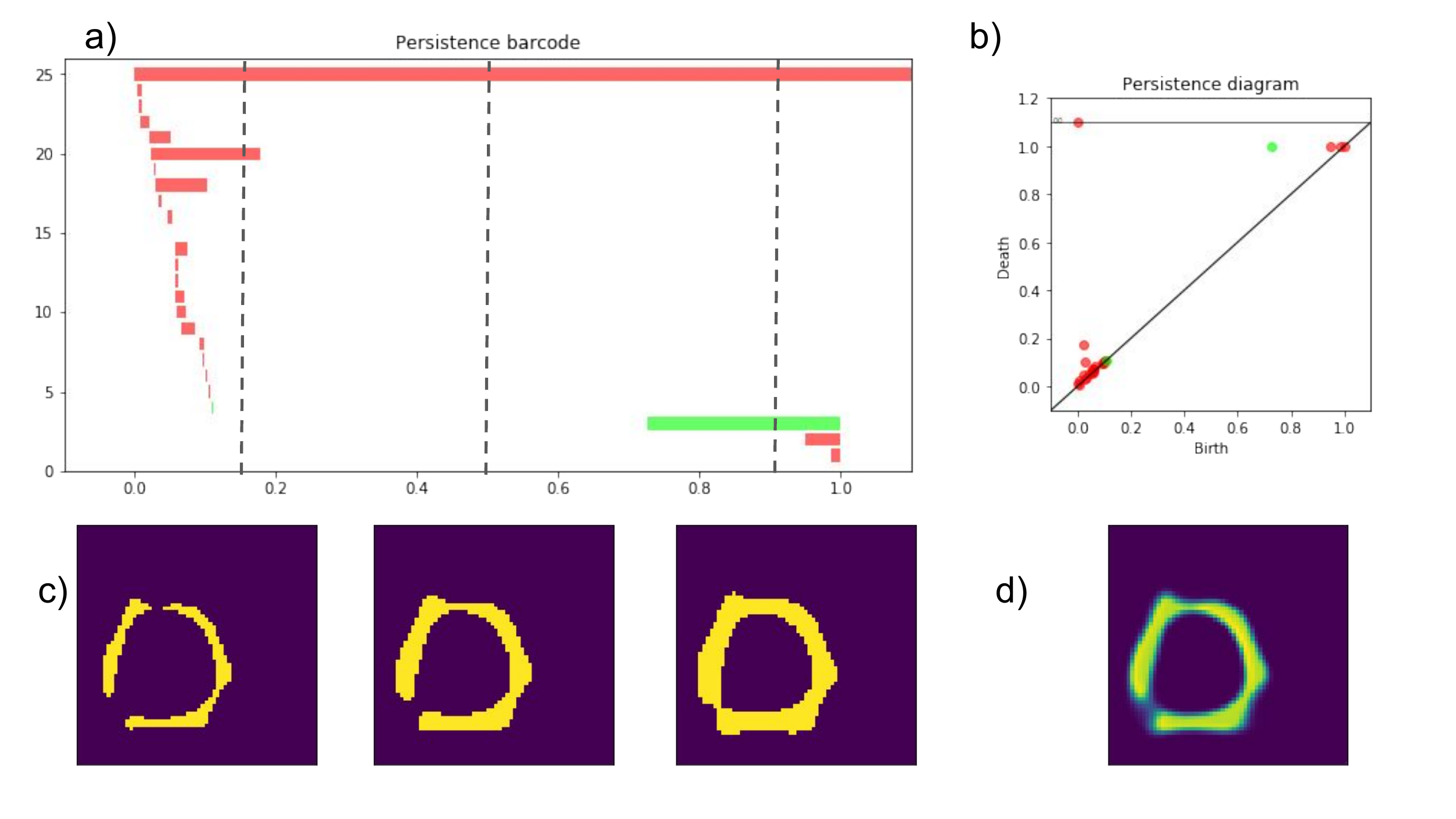}}
    \caption{The barcode diagram for the predicted mask shown in d).
    As the filtration value is increased, moving left to right on the barcode diagram a), more pixels join the cubical complex, and topological features are created and destroyed.
    0-dimensional features (i.e. connected components) are shown by red bars, and 1-dimensional features (i.e. closed cycles) by green bars.
    The pixels in the cubical complex for three filtration values (corresponding to the vertical dashed lines on the barcode diagram) are shown here in c).
    In the first complex, there are two connected components, corresponding to the two red bars crossing the first dashed line.
    In the second, there is only one connected component (and so one red bar).
    In the third, one connected component and one closed cycle (and so one red, and one green bar) are present.
    The persistence diagram showing the lifespans of these topological features is shown in b).}
    \label{fig:barcode_example}
\end{figure}

A useful way of visualising the PH of a dataset is to use a \emph{barcode diagram}, an example of which is given in Figure \ref{fig:barcode_example}.
This diagram plots the lifespans of all topological features in the data, where each feature is represented by one bar, and with different colour bars representing different Betti numbers.
The Betti numbers of $B(p^*)$ are given by the number of bars present at the x-coordinate $x=p^*$.
A key feature of the barcode diagram is that it is stable in the presence of noise, and there are theoretical guarantees that small changes to the original data %$S$
can only make small changes to the positions and lengths of the bars \cite{Ghrist2008}.
For an array of input data $S$, we will describe its PH by denoting each bar in the barcode diagram as $H^\ell_d(S)=(p_{\mathrm{birth}}, p_{\mathrm{death}})$ which is an ordered pair of the birth and death filtration values of the $\ell^\mathrm{th}$ longest bar of dimension $d$, where $\ell \geq 1 $ and $d \geq 0$.

% JC: don't need to discuss persistence diagram
%Another commonly used representation is the persistence diagram, an example of which is also shown in figure \ref{fig:barcode_example}, in which each topological feature is plotted with its birth and death filtration values on the x-axis and y-axis respectively.
%In this diagram, the distance from the diagonal indicates the lifespan of the feature.

Our method will use these barcode diagrams as a description of the topological features in a predicted segmentation mask.
In the case we consider below, we begin with the prior knowledge that the object being segmented should contain one hole (i.e. $\beta_1=1$) and so aim to extend the length of the bar corresponding to the most persistent 1-dimensional feature.
It is important to note both that our method can be applied generally to encourage the presence or absence of topological features of any number or dimension, but also that this prior information must be specified for the task at hand.

\section{Method}
\label{sec:method}
Throughout this paper we consider only the problem of binary segmentation, that is, assigning a value between $0$ and $1$ to each pixel in an image which represents the probability of it being classified as part of a particular structure.
Our approach does generalise to multi-class segmentation (inasmuch as it can be described as several binary segmentation problems) but, for convenience and simplicity, we will discuss only the binary case here.

\subsection{Topological Pixelwise Gradient}
In our approach the desired topology of the segmentation mask needs to be specified in the form of its Betti numbers.
For ease of explanation we consider the case in which $\beta_1=1$ is specified, corresponding to the prior knowledge that the segmentation mask should contain exactly one closed cycle.
Given a neural network $f$ which performs binary segmentation and is parameterised by a set of weights $\omega$, an $N \times N$ image $X$ produces an $N \times N$ array of pixelwise probabilities, $S=f(X; \omega)$.
In the supervised learning setting, a pixelwise gradient is calculated by, for example, calculating the binary cross-entropy or Dice loss between $S$ and some ground-truth labels $Y$.

We additionally calculate a pixelwise gradient for a topological loss as follows.
Firstly the PH of $S$ 
%(not of the original image $X$) %
is calculated, producing a set of lifetimes of topological features $H_d^\ell$, such as that shown in figure \ref{fig:barcode_example}a.
For each desired feature, the longest bars (and so the most persistent features) of the corresponding dimension are identified.
In our case the presence of a closed cycle corresponds to the longest green bar in the barcode diagram, denoted by $H_1^1(S)$.
In order to make this feature more persistent we need to identify the pixels which, if assigned a higher/lower probability of appearing in the segmentation, will extend the length of this bar in the barcode diagram, and therefore increase the persistence of that topological feature.
%
% JC: This algorithm should describe extending the lifetime of the bar in both directions
%     But there is a threshold value, say, 0.01 around which we don't apply a gradient
%     And in our examples that will apply to the upper end of the bar
%     
These pixels are identified by an iterative process which begins at the pixels with the filtration values at precisely the ends of the relevant bar, which are $H_1^1(S)=(p^*_\mathrm{birth}, p^*_\mathrm{death})$ for the left and right ends of the bar respectively.
For each of $k$ iterations, where $k$ is an integer parameter which can be freely chosen, the pixels with these extremal filtration values are filled in (with a $1$ and $0$ respectively), extending the bar in the barcode, and these pixels have a gradient of $\mp 1$  applied to them.
The PH is recomputed, and another pixel chosen for each end of the bar.
These pixels are also filled in, and more chosen, and so on, until $k$ pixels have been identified for each end of the bar.
These are now the $2k$ pixels which, if their filtration values are adjusted, will result in the most significant change in the persistence of the relevant topological object, and it is these $2k$ pixels which will have a gradient applied to them.
Algorithm \ref{alg:topograd} shows pseudo-code for the $\beta_1=1$ example%
\footnote{Our implementation will be made publicly available upon publication.}%
.

\begin{algorithm}[ht]
\caption{Topological loss gradient $\beta_1=1$}\label{alg:topograd}
\hspace*{\algorithmicindent} \textbf{Input} \\
\hspace*{\algorithmicindent} $S$: Array of real numbers - pixelwise segmentation probabilities \\
\hspace*{\algorithmicindent} $k$: Integer - number of pixels to apply gradient to \\
\hspace*{\algorithmicindent} $\epsilon$: Real number $> 0$ - threshold to avoid modifying already persistent features \\
\hspace*{\algorithmicindent} \textbf{Output} \\
\hspace*{\algorithmicindent} $G$: Array of real numbers - pixelwise gradients
\begin{algorithmic}[1]
\Procedure{TopoGrad}{$S,k,\epsilon$}
\State $t \gets 0$
\State $G$ is initialised as an $N \times N$ array of $0$
\While{$t < k$}
\State $H_d^\ell(S) \gets \mathrm{PersistentHomology}(S)$\Comment{Calculate Barcode Diagram of S}
\State $(p^*_\mathrm{birth}, p^*_\mathrm{death}) \gets H_1^1(S)$\Comment{Longest bar for 1-dimensional features}
\For{pixel [i,j] in array $S$}
\If{$p^*_\mathrm{birth} > \epsilon$}\Comment{Skip if bar already begins close to $p=0$}
\If{$S_{[i,j]}=1-p^*_\mathrm{birth}$}
        \State $S_{[i,j]} \gets 1$ \Comment{Set the pixel with value $1-p^*_\mathrm{birth}$ to 1}
        \State $G_{[i,j]} \gets -1$ \Comment{Set its gradient to $-1$}
      \EndIf
  \EndIf
\If{$p^*_\mathrm{death} < (1-\epsilon)$}\Comment{Skip if bar already ends close to $p=1$}
\If{$S_{[i,j]}=1-p^*_\mathrm{death}$}
        \State $S_{[i,j]} \gets 0$ \Comment{Set the pixel with value $1-p^*_\mathrm{death}$ to 0}
        \State $G_{[i,j]} \gets +1$ \Comment{Set its gradient to $+1$}
      \EndIf
  \EndIf
      \EndFor
\State $t \gets t+1$
\EndWhile\label{topograd_while}
\State \textbf{return} $G$\Comment{Return the pixelwise gradient array}
\EndProcedure
\end{algorithmic}
\end{algorithm}

\subsection{Semi-supervised Learning}
\label{sec:ssl}
We incorporate the topological prior into a semi-supervised learning scheme as follows.
In each training batch, firstly the binary cross-entropy loss from the $N_\ell$ labelled cases is calculated.
Next the pixelwise gradients, $G$, for the $N_u$ unlabelled cases are calculated as in Algorithm \ref{alg:topograd} and multiplied by a positive constant $\lambda$, which weights this term.
The gradient from the cross-entropy loss and the topological gradient are then summed.
In our experiments we set $k=5$, $\epsilon=0.01$ and experiment with a choice of $\lambda$, chosen by manual tuning.

\section{Experiments and Results}
\label{sec:application}
\begin{figure}[ht]%
\vspace{-5mm}
\noindent\makebox[\textwidth]{
    \centering
    \subfloat[]{{\includegraphics[width=0.205\textwidth]{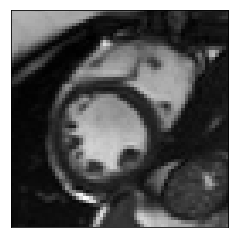}}}%
    \subfloat[]{{\includegraphics[width=0.205\textwidth]{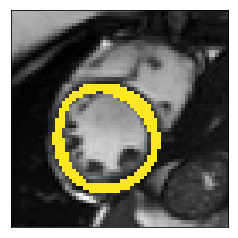}}}%
    \subfloat[]{{\includegraphics[width=0.205\textwidth]{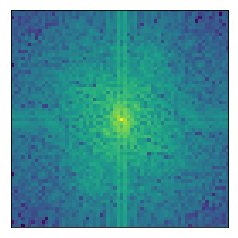}}}%
    \subfloat[]{{\includegraphics[width=0.205\textwidth]{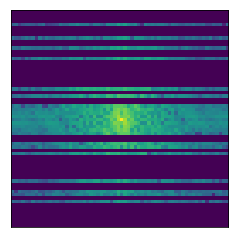}}}%
    \subfloat[]{{\includegraphics[width=0.205\textwidth]{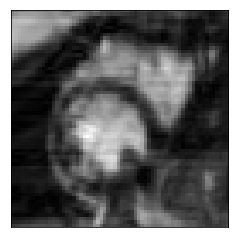}}}}%
    \caption{An example image (a) and segmentation (b) from the UK Biobank dataset.
    We degrade images by transforming them to k-space (c) and randomly removing non-central k-space lines (d), making the task of myocardial segmentation more challenging by causing blurring and aliasing artefacts (e).}%
    \label{fig:degradation}%
\end{figure}

We demonstrate our approach on real data with the task of myocardial segmentation of cardiac MRI.
We use a subset of the UK Biobank dataset \cite{Sudlow2015,Petersen2016}, which consists of the mid-slice of the short-axis view of the heart.
Example images and segmentations from this dataset are shown in Figure %\ref{fig:ukbb_examples}.
\ref{fig:degradation}.
%
\iffalse
\begin{figure}[ht]
    \centering
    \includegraphics[width=\textwidth]{ukbb_examples_3.png}
    \caption{Example images and corresponding segmentation masks from the UK Biobank dataset.
    Images and segmentations are cropped to a 64x64 region of interest centred around the left ventricle.}
    \label{fig:ukbb_examples}
\end{figure}
\fi
%
We use one end-systole image from each subject, each of which has a gold-standard left-ventricle segmentation provided.
The images were cropped to a 64x64 square centred around the left ventricle.
Since the UK Biobank dataset contains high-quality images compared to a typical clinical acquisition we made the task more challenging,  degrading the images by removing k-space lines in order to lower image quality and create artefacts.
For each image in the dataset we compute the Fourier transform, and k-space lines outside of a central band of $8$ lines are removed with $3/4$ probability and zero-filled.
The degraded image is then reconstructed by performing the inverse Fourier transform, and it is these images which are used for both training and testing.
Examples of original and degraded images are shown in Figure \ref{fig:degradation}.
Our method is demonstrated in the semi-supervised setting, where a small number of labelled cases, $N_\ell$, and $N_u=400$ unlabelled cases are used. 
As a baseline we evaluated a fully supervised method using just the labelled cases, and also post-processed the supervised results using image processing tools commonly used to correct small topological errors.
We used the binary closure morphology operator with a circular structuring element with a radius of 3 pixels.
Additionally, we compared our method to an iterative semi-supervised approach similar to \cite{Bai2017}.
In this method the predicted segmentations from unlabelled cases were used as labels for training such that the network's weights and the predicted segmentations of unlabelled cases are iteratively improved.
In our experiments, as in \cite{Bai2017} we use 3 iterations of 100 epochs after the initial supervised training.

Each of these methods was evaluated with the same network architecture.
We used a simple U-net-like network \cite{Ronneberger2015} but with 3 levels of spatial resolution (with 16, 32, and 64 feature maps in each, and spatial downsampling by a factor of 2) and with 3 3x3  convolution plus ReLU operations before each upsampling or downsampling step, with the final layer having an additional 1x1 convolution followed by a sigmoidal activation.
This results in 16 convolutional layers in total.
All models were trained using the Adam optimiser with a learning rate of $10^{-4}$ and the supervised part of the model was trained with the Dice loss.
The trained networks were then evaluated against a held-out test set of $N_{\mathrm{test}}=500$ cases.
To evaluate our approach we measured the Dice score of the predicted segmentations, as a quantifier of their pixelwise accuracy, and the proportion of segmentations with the correct topology when thresholded at $p=0.5$.

Table \ref{tab:results} shows the mean results over $500$ test cases averaged over $20$ training runs (over which both the allocation of images into training and test sets and the image degradation were randomised).
Our method provides a significant reduction in the proportion of incorrect topologies of the segmentations compared to the baseline supervised learning scenario.
Notably, this can occur without significantly sacrificing the pixelwise metrics of segmentation quality demonstrating that an increase in topological accuracy does not need to come at a cost to pixelwise accuracy.
In Figure \ref{fig:clinical_exemplar} we show a typical clinically acquired short-axis image and its estimated segmentations with and without our method.
This image has not been artificially degraded as in our experiment above and is shown to illustrate that clinically acquired scans are often of a low quality compared to the UK Biobank dataset on which we demonstrate our method, and so are challenging to segment.
Qualitatively observing these cases we see that a topological prior is beneficial in this realistic scenario.
\begin{table}[ht]
    \centering
\begin{tabular}{l|r r r r r | r r r r r}
 %& \multicolumn{2}{c}{$N_s=20$} & \multicolumn{2}{c}{$N_s=50$} & \multicolumn{2}{c}{$N_s=100$} & \multicolumn{2}{c}{$N_s=200$} & \multicolumn{2}{c}{$N_s=500$} \\
 & \multicolumn{5}{c}{Dice Score} & \multicolumn{5}{c}{Percentage of correct topologies} \\
 $N_\ell$ & 20 & 40 & 100 & 200 & 400  & 20 & 40 & 100 & 200 & 400 \\
 \toprule
SL         & \, $0.739$ & \, $0.762$ & \, $0.802$ & \, $0.824$ & \, $0.842$ & \, $69.1\%$ & \, $78.1\%$ & \, $89.0\%$ & \, $92.8\%$ & \, $96.2\% \,$\\
SL + BC    &    $0.739$ &    $0.762$ &    $0.802$ &    $0.823$ &    $0.841$ &    $70.2\%$ &    $78.9\%$ &    $89.3\%$ &    $92.8\%$ & \, $95.6\%$ \\
SSL  & \,       $0.740$ & \, $0.763$& \,  $0.801$ & \, $0.826$ & \, $0.842$ & \, $71.4\%$ & \, $79.2\%$ & \, $89.7\%$ & \, $93.6\%$ & \, $95.4\%$\\
Ours $\lambda=1$ & $0.742$ & $0.762$ &    $0.803$ &    $0.825$ & $0.843$ & $\mathbf{77.4\%}$ & $\mathbf{84.4\%}$ & $\mathbf{91.8\%}$ & $\mathbf{94.0\%}$ & $\mathbf{96.4\%}$  \\
Ours $\lambda=3$ & $0.730^*$ & $0.749^*$ &    $0.792$ &    $0.824$ &     $0.844$ & $\mathbf{82.9\%}$ & $\mathbf{89.8\%}$ & $\mathbf{95.5\%}$ & $\mathbf{94.9\%}$ & $\mathbf{96.7\%}$  
\vspace*{3mm}
\end{tabular}
    \caption{Comparison of supervised learning (SL), supervised learning with binary closure (SL + BC), the semi-supervised approach of 
    \cite{Bai2017}
    %Bai et. al 
    (SSL) and our semi-supervised learning (Ours) with a topological prior, averaged over the 20 training/testing runs.
    Bolded results are those which are statistically significantly better than all three benchmark methods.
    Starred results are statistically significantly worse than at least one benchmark method.
    Significance was determined by $p<0.01$ when tested with a paired t-test.
    }
    \label{tab:results}
\vspace*{-3mm}
\end{table}

\section{Discussion}
\label{sec:discussion}
Although we have only demonstrated our approach for the segmentation of 2D images here, in the often challenging task of 3D segmentation, the ability to impose a topological loss function could be of significant use as the number of connected components, handles, and cavities may be specified.
Our future work will investigate this generalisation and its utility in challenging tasks such as the 3D segmentation of cardiac MRI volumes of subjects with congenital conditions causing atypical connections between chambers of the heart.
We will also investigate extending our approach to incorporate first learning the topology of a structure from the image, and then incorporating that knowledge into the segmentation, which would allow our approach to be applicable to cases such as cell segmentation where the number of components in the desired segmentation is not known \emph{a priori} but can be deduced from the image. 

%Although in our experiments (see table \ref{tab:results}) our method does not have a statistically significantly worse Dice score than other methods, we have observed in this and other informal experiments that the pixelwise accuracy sometimes decreases slightly with our method.
%In our experiments we considered only one value of the weight of the topological prior, $\lambda$, in detail, but we have found that varying this quantity effectively allows one to trade off the pixelwise and topological accuracies.
%When $\lambda$ is small, there is little change from the original segmentation, and when it is large the algorithm heavily over-segments which ensures topological correctness but at significant cost to pixelwise accuracy.
%In future work we will also investigate the extent to which this trade-off can be managed so as to learn the optimal value of $\lambda$ for a given objective.

In our experiments we found that setting $\lambda=1$ meant that our method had no significant difference in Dice score to the other methods but an improved topological accuracy.
As seen in table \ref{tab:results} a higher $\lambda$ results in even better performance according to the segmentation topology, but pixelwise accuracy begins to drop.
We found that changing $\lambda$ allows one to trade off the pixelwise and topological accuracies and in future work we will also investigate the extent to which this trade-off can be managed so as to learn the optimal value of $\lambda$ for a given objective.

The dominant computational cost in our method is the repeated PH calculation which occurs $k$ times when calculating the pixelwise gradients for each image.
Computing the PH for a cubical complex containing $V$ pixels/voxels in $d$ dimensions can be achieved in $\Theta(3^d V + d 2^d V)$ time, and $\Theta(d 2^d V)$ memory (see \cite{Wagner2012}) and so scales linearly with respect to the number of pixels/voxels in an image.
%, with 3D volumes having $3$ times the cost of 2D images.
We found that, using $64 \times 64$ pixel 2D images, the PH for one image was calculated in approximately $0.01$s on a desktop PC.
Consequently, when using $k=5$ and a batch of $100$ images for semi-supervised learning, one batch took about $5$s to process.
On large 3D volumes this cost could become prohibitive.
However, the implementation of PH that we use is not optimised for our task and our algorithm allows for parallel computation of the PH of each predicted segmentation in the batch of semi-supervised images.
With a GPU implementation for calculating the PH of a cubical complex, many parallel calculations could allow for significant improvements in overall run-time.

\section{Conclusions}
\label{sec:conclusion}
We have presented the first work to incorporate explicit topological priors into deep-learning based image segmentation, and demonstrated our approach in the 2D case using cardiac MRI data.
We found that including prior information about the segmentation topology in a semi-supervised setting improved performance in terms of topological correctness on a challenging segmentation task with small amounts of labelled data.

 \bibliographystyle{splncs04}

\end{document}